\documentclass[10pt,twocolumn,letterpaper]{article}
\usepackage[final]{cvpr} 

\usepackage{times}
\usepackage{epsfig}
\usepackage{graphicx}
\usepackage{amsmath}
\usepackage{amssymb}
\usepackage{chngpage}
\usepackage{mathtools}
\usepackage{caption}
\usepackage[table,xcdraw]{xcolor}
\usepackage[section]{placeins}
\usepackage{multirow}
\usepackage{listings}

\usepackage[pagebackref=true,breaklinks=true,colorlinks,bookmarks=false]{hyperref}

\def\cvprPaperID{13} % *** Enter the CVPR Paper ID here
\def\confName{CVPR}
\def\confYear{2022}

\begin{document}

%%%%%%%%% TITLE
\title{X-ViT: High Performance Linear Vision Transformer without Softmax}

\author{
Jeonggeun Song\thanks{Equal contribution}\, $^\dag$, Heung-Chang Lee$^\ast$\thanks{Corresponding authors}\\
Kakao Enterprise\\
Sengnam-si, Republic of Korea\\
{\tt\small po.ai; andrew.com@kakaoenterprise.com}
% \and
}
\maketitle

\maketitle
\begin{abstract}
Vision transformers have become one of the most important models for computer vision tasks. Although they outperform prior works, they require heavy computational resources on a scale that is quadratic to the number of tokens, $N$. This is a major drawback of the traditional self-attention (SA) algorithm. Here, we propose the X-ViT, ViT with a novel SA mechanism that has linear complexity. The main approach of this work is to eliminate nonlinearity from the original SA. We factorize the matrix multiplication of the SA mechanism without complicated linear approximation. By modifying only a few lines of code from the original SA, the proposed models outperform most transformer-based models on image classification and dense prediction tasks on most capacity regimes.
\end{abstract}

\section{Introduction}
As early successes in natural language processing (NLP), several studies based on
transformers %\cite{dosovitskiy2020image, touvron2020training, wu2021cvt, srinivas2021bottleneck, heo2021rethinking, graham2021levit, el2021xcit}
have shown impressive results in vision tasks. Recent studies have shown that transformer-based architectures renew the state of the art across a wide range of subject areas, including image classification%\cite{dosovitskiy2020image, touvron2020training}
, object detection and semantic segmentation%\cite{carion2020end, liu2021swin, wang2021pyramid, zhang2021multi, zheng2021rethinking}
, and generative models.%\cite{jiaeng2021transgan, esser2021taming, durall2021combining}.

Despite its great successes, the original self-attention (SA) mechanism has $O(N^2)$ time and memory complexity due to the matrix multiplication of $\sigma (QK^T)\in R^{N\times N}$ and $V$. This is one of the well-known drawbacks of traditional transformers. For vision tasks, $N$ is proportional to the input resolution. This means that SA consumes 16 times the computational resources if the width and height of the input image are doubled.

Here, we propose a new model that implements an alternative novel SA mechanism to avoid this drawback. It is called the X-ViT, the vision transformer with XNorm. Our key method replaces softmax nonlinearity with a simple $L_2$-norm. Using the associative law of matrix multiplication, our new SA algorithm requires much less computational resources than the original SA.
\begin{figure}[t!]
\begin{center}
\includegraphics[width=\linewidth]{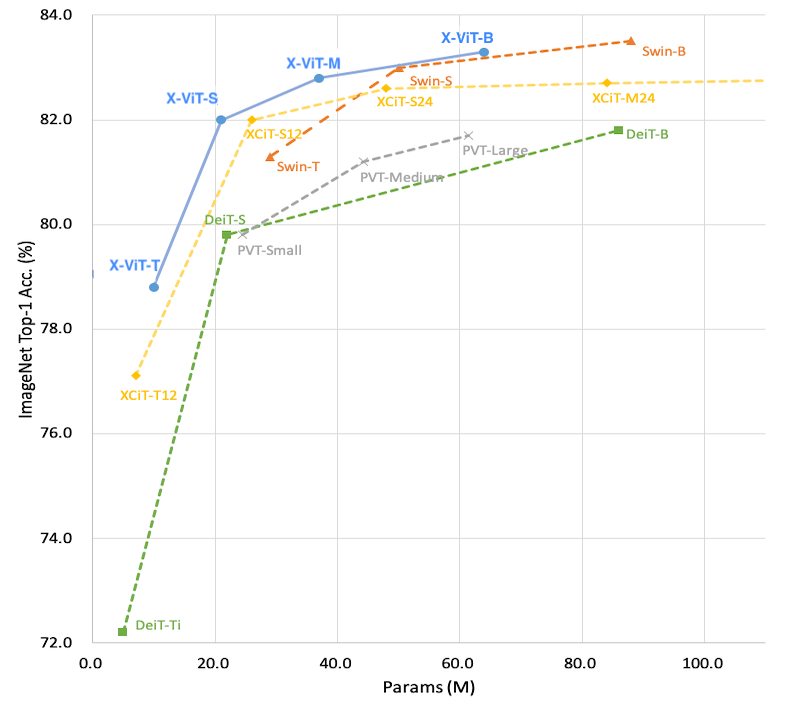}
\end{center}
\vskip -0.05in
\caption{\textbf{Top-1 accuracy vs. Model capacity.} Comparison of ImageNet1k top-1 accuracy of various models according to model capacity. Our models show the best results at the same parameter sizes compared to the other models.}
\label{fig:params_acc_comparison}
\vskip -0.05in
\end{figure}

%Although similar approaches have been proposed by Performer\cite{heo2021rethinking} and Efficient Attention\cite{shen2021efficient}, El et al. have shown that these approaches cause performance degradation for XCiT\cite{el2021xcit}. Unlike the previous results, X-ViT achieves higher or competitive results on the benchmark datasets of vision tasks compared with state-of-the-art models. This is detailed in Section \ref{sec:experiments}, including the results on the ImageNet1k\cite{deng2009imagenet} and COCO benchmarks\cite{lin2014microsoft}. These results were obtained without any extra-large datasets, such as ImageNet21k, or distilled knowledge from another model.

The main contributions in this paper are summarized as follows:
\begin{itemize}
    \item We propose a novel \textit{constraint} scheme, XNorm, that generates a unit hypersphere to extract relational features. %As detailed in Section \ref{sec:methods}, this scheme prevents SA from being dependent on initialization. Furthermore, i
    It eliminates non-linearity from SA by replacing the softmax function. Our module has $\text{O}(N)$ complexity, and it handles high-resolution inputs efficiently.
    \item We demonstrate that X-ViT can be adopted for general purposes. %Our models are tested on both image classification and dense prediction tasks. While our proposed method has linear complexity, the X-ViT models 
    Our proposed method outperforms most of the state-of-the-art models based on transformers at lower capacity and FLOPs. In particular, our models perform well in lightweight regimes.
    \item We empirically show that X-ViT models have faster inference speed and require less GPU memory. %As the input resolution increased, the required computing resources increased less than the other models.
\end{itemize}

\section{Related Works}
Dosovitskiy et al.\cite{dosovitskiy2020image} proposed a
vision transformer (ViT), which showed that transformer-based models could be used for vision tasks. After the achievements of ViT, DeiT\cite{touvron2020training} introduced data-efficient training strategies for vision transformers with detailed ablation studies. They solved the ViT data efficiency problem successfully, and most of the current transformer-based models follow their schemes.

Instead of architectural strategies, many approaches have been proposed to solve the $\text{O}(N^2)$ problem of the SA mechanism. They are summarized in several categories: those that use their own spatial patterns\cite{ho2019axial, child2019generating, sukhbaatar2019adaptive}, those that use various low-rank factorization methods \cite{choromanski2020rethinking, shen2021efficient, wang2020linformer}, those that use linear approximation by sampling important tokens\cite{kitaev2020reformer, xiong2021nystr}, and those that use cross-covariance matrices instead of Gram matrices\cite{el2021xcit}. Although detailed methods are quite different, our XNorm is mainly related to low-rank factorization methods.

Tokens-to-token ViT, introduced by Yuan et al.\cite{yuan2021tokens}, aims to achieve a similar objective through different approaches. They presented a method of overlapping tokens to locally correlate patches. They did not use additional methods to reduce the computation, except when using small channels. El-Nouby et al. introduced local patch interactions in XCiT\cite{el2021xcit}. With two depthwise convolutions\cite{chollet2017xception} added after XCA, XCiT achieved better performance. Our models are generally inspired by the intrinsic optimization strategies that XCiT\cite{el2021xcit} introduced, while we present our own SA method.

\section{Methods} \label{sec:methods}
\begin{figure}[t] % t: top, b: bottom, h: here
\begin{center}
\includegraphics[width=0.7\linewidth]{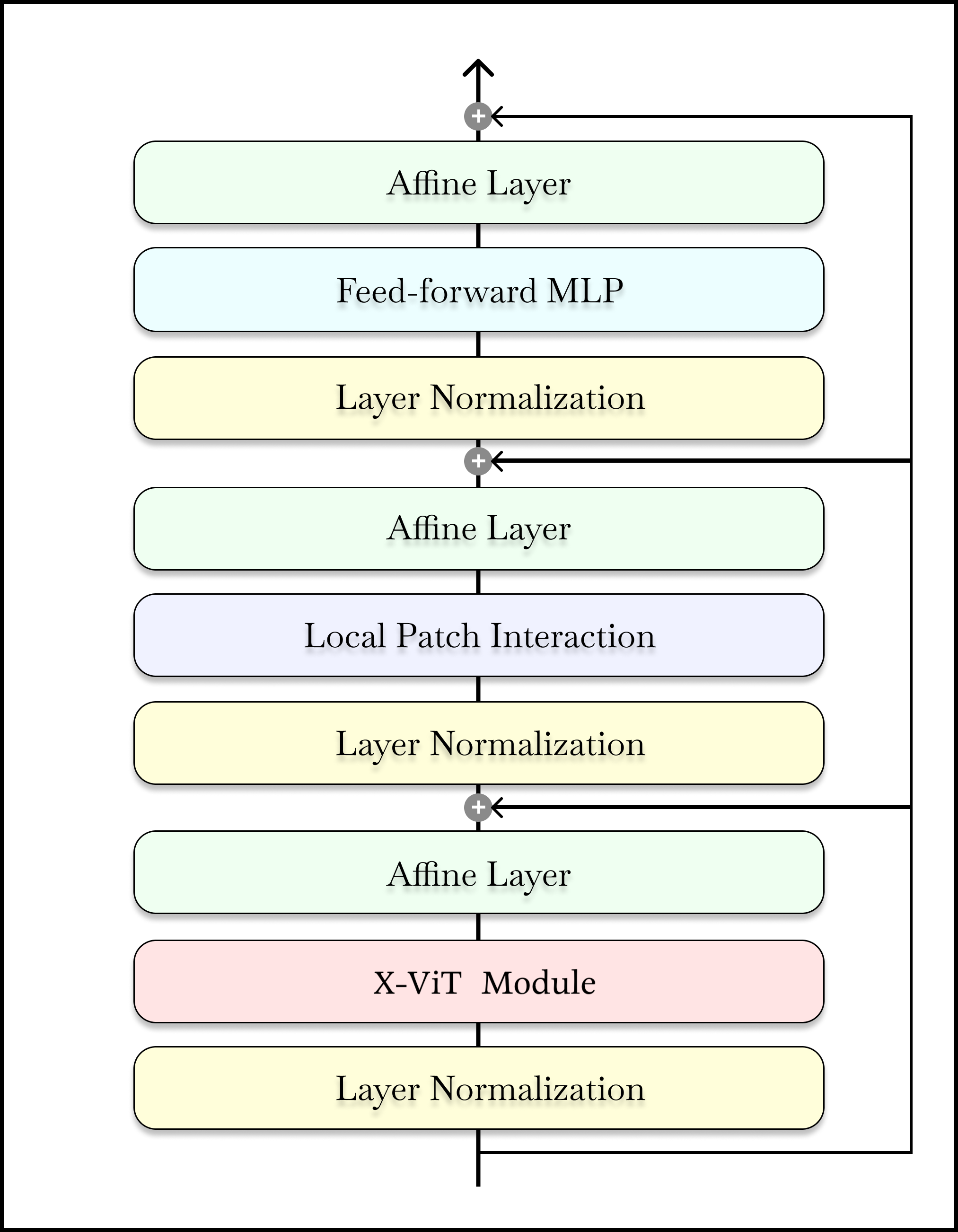}
\end{center}
\caption{\textbf{Overview of X-ViT module.} Note that affine layers\cite{touvron2021resmlp} are following each module.} \label{fig:ufo-vit}
\end{figure}

\begin{figure}[t] % t: top, b: bottom, h: here
\begin{center}
\includegraphics[width=\linewidth]{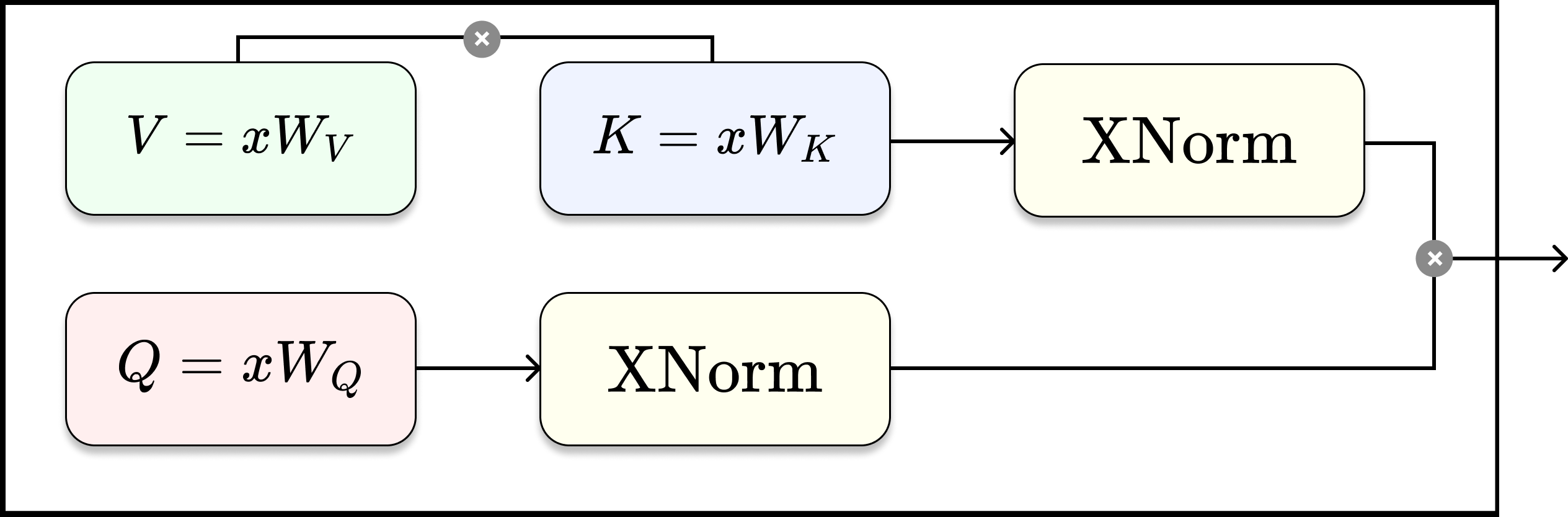}
\end{center}
\caption{\textbf{X-ViT module.}} \label{fig:ufo_module}
\end{figure}
\subsection{XNorm}
The structure of our model is shown in Figure \ref{fig:ufo-vit}. It is a mixture of convolutional layers, a X-ViT module, and a simple feed-forward MLP layer.

For an input $\textbf{x} \in \mathbb{R}^{N\times C}$, the original SA mechanism is formulated as follows:
\begin{align}\label{eq:sa}
&A(\textbf{x})=\text{Softmax}(QK^T/\sqrt{d_k})V\\
&Q=\textbf{x}W_Q, K=\textbf{x}W_K, V=\textbf{x}W_V
\end{align}
where $A$ denotes the attention operator.

By removing softmax function from original SA, $QK^{T}V$ can be decomposed into $K^T$ and $Q(K^{T}V)$. Compared to $\text{O}(N^2)$ complexity of original SA, each matrix multiplication has $\text{O}(N)$ complexity.

So we designed a simple constraint to replace softmax function. Our proposed method, called \textbf{cross-normalization} or \textbf{XNorm}, is defined as follows:
\begin{align}\label{eq:xnorm} &A(\textbf{x})=\text{XN}_{\text{dim}=\text{filter}}(Q)(\text{XN}_{\text{dim}=\text{space}}(K^{T}V))\\
&\text{XN}(\textbf{a})\coloneqq \cfrac{\gamma\textbf{a}}{\sqrt{\sum_{i=0}^{h} ||\textbf{a}||^2}}
\end{align}
where $\gamma$ is a learnable parameter and $h$ is the number of embedding dimensions. It is a common $L_2$-norm, applied to the patches of $K^{T}V$ and the filters of $Q$.

In the above formulation, the patches of $V$ are projected to $h$ dimension by $K$. After that, the pixel-to-pixel relations are computed by multiplying $Q$. In this process, we observed that the variance of sizes of the pixel vectors can harm the stability of training at initial stage. With XNorm, all pixels are normalized to unit-sized vectors. It makes training stable and improves the performance of the model.

To build our X-ViT model, we adopted architectural strategies from earlier vision transformer models\cite{graham2021levit, xiao2021early, el2021xcit, touvron2021going}. First, we used convolutional layers instead of linear patch embedding layers. Several recent studies\cite{graham2021levit, xiao2021early} claimed that early convolutional layers help vision transformers to be well-trained. Also, we added the local patch interaction (LPI) layers proposed in XCiT\cite{el2021xcit}. We found that the latter showed better performance than the other type of convolutional modules. The overall structure is illustrated in Figure \ref{fig:ufo-vit}.

\begin{table}[t!]
\centering
\begin{tabular}{l|c|c|c}
\hline
\rowcolor[HTML]{EFEFEF}
Model & Top-1 Acc. & Params & FLOPs \\ \hline
RegNetY-1.6G\cite{radosavovic2020designing} & 78.0 & 11M & 1.6G \\
DeiT-Ti\cite{touvron2020training} & 72.2 & 5M & 1.3G \\
XCiT-T12/16\cite{el2021xcit} & 77.1 & 26M & 1.2G \\
\rowcolor[HTML]{EFEFEF}
X-ViT-T & 78.8 & 10M & 1.9G \\ \hline

ResNet-50\cite{he2016deep} & 75.3 & 26M & 3.8G \\
RegNetY-4G\cite{radosavovic2020designing} & 80.0 & 21M & 4.0G \\
DeiT-S\cite{touvron2020training} & 79.8 & 22M & 4.6G \\
Swin-T\cite{liu2021swin} & 81.3 & 29M & 4.5G \\
XCiT-S12/16\cite{el2021xcit} & 82.0 & 26M & 4.8G \\
\rowcolor[HTML]{EFEFEF}
X-ViT-S & 82.0 & 21M & 3.7G \\ \hline

ResNet-101\cite{he2016deep} & 75.3 & 47M & 7.6G \\
RegNetY-8G\cite{radosavovic2020designing} & 81.7 & 39M & 8.0G \\
Swin-S\cite{liu2021swin} & 83.0 & 50M & 8.7G \\
XCiT-S24/16\cite{el2021xcit} & 82.6  & 48M & 9.1G \\
\rowcolor[HTML]{EFEFEF}
X-ViT-M & 82.8 & 37M & 7.0G \\ \hline

RegNetY-16G\cite{radosavovic2020designing} & 82.9 & 84M & 16.0G \\
DeiT-B\cite{touvron2020training} & 81.8 & 86M & 17.5G \\
Swin-B\cite{liu2021swin} & 83.5 & 88M & 15.4G \\
XCiT-M24/16\cite{el2021xcit} & 82.9 & 84M & 16.2G \\
\rowcolor[HTML]{EFEFEF}
X-ViT-B & 83.3 & 64M & 11.9G \\ \hline

EfficientNet-B7\cite{tan2019efficientnet} & 84.3 & 66M & 37.0G \\
XCiT-S24/8\cite{el2021xcit} & 83.9 & 48M & 36.0G \\
Swin-B/384\cite{liu2021swin} & 84.5 & 48M & 47.0G \\
\rowcolor[HTML]{EFEFEF}
X-ViT-M/384 & 83.8 & 37M & 20.5G \\
\rowcolor[HTML]{EFEFEF}
X-ViT-B/384 & 84.3 & 64M & 35.1G \\ \hline
\end{tabular}
\caption{\textbf{Comparison with the state of the art models.} The image classification results, model capacity, and FLOPs of various models on ImageNet1k dataset.} \label{tb:imagenet_comparison}
\end{table}

\subsection{X-ViT}\label{sec:x-vit}
To build our X-ViT model, we adopted architectural strategies from earlier vision transformer models\cite{graham2021levit, xiao2021early, el2021xcit, touvron2021going}. In this section, we introduce several intrinsic structures that improve performance. The overall structure is illustrated in Figure \ref{fig:ufo-vit}.

\textbf{Replace linear patch embedding with convolutions.} Several recent studies\cite{graham2021levit, xiao2021early} claimed that early convolutional layers help vision transformers to be well-trained. To adopt their strategy, we used convolutional layers instead of linear patch-embedding layers.

\textbf{Multi-headed attention.} Following the original transformer\cite{vaswani2017attention}, our modules are multi-headed for better regularization. The $\gamma$ parameter in Eq.\ref{eq:xnorm} is applied to all heads to scale the importance of each head.

\textbf{Convolutional layers.} Designing an extra module to extract local features is not a new idea. We chose the most simplistic method by adding various types of convolutional layers. We experimented with both the simple depthwise convolutions and the local patch interaction (LPI) layers proposed in XCiT\cite{el2021xcit}. We found that the latter showed better performance on the regimes overall.

\textbf{Class attention.} In the ImageNet1k experiments, we used the class attention layers presented in CaiT\cite{touvron2021going}. This helps the class token gather spatial information. Class attention is computed on class token only to reduce computation, as in the original paper. We implemented the class attention layers using X-ViT modules, whereas CaiT used the SA module for class attention.

\begin{table}[t!]
\centering
\begin{tabular}{l|c|c|c}
\hline
\rowcolor[HTML]{EFEFEF} 
Backbone &
  Params &
  $\text{mAP}^b$ &
%   $\text{AP}^b_{50}$ &
%   $\text{AP}^b_{75}$ &
  $\text{mAP}^m$ \\ \hline
%   $\text{AP}^m_{50}$ &
%   $\text{AP}^m_{75}$ \\ \hline
ResNet50\cite{he2016deep} & 44M & 41.0 & 37.1 \\
PVT-Small\cite{wang2021pyramid} & 44M  & 43.0 & 39.9 \\
Swin-T\cite{liu2021swin} & 48M & 46.0 & 41.6 \\
XCiT-S12/16\cite{el2021xcit} & 44M  & 45.3 & 40.8 \\
\rowcolor[HTML]{EFEFEF} 
X-ViT-S & 40M  & 44.6 & 40.4 \\ \hline
ResNet101\cite{he2016deep} & 63M  & 42.8 & 39.2 \\
PVT-Medium\cite{wang2021pyramid} & 64M  & 44.2 & 40.5 \\
Swin-S\cite{liu2021swin} & 69M & 48.5 & 43.3 \\
XCiT-S24/16\cite{el2021xcit} & 66M  & 46.5 & 41.8 \\
\rowcolor[HTML]{EFEFEF} 
X-ViT-M & 56M & 46.0 & 41.0\\ \hline
ResNeXt101-64\cite{xie2017aggregated} & 102M  & 44.4 & 39.7 \\
PVT-Large\cite{wang2021pyramid} & 81M  & 44.5 & 40.7 \\
XCiT-M24/16\cite{el2021xcit} & 101M & 46.7 & 42.0 \\
\rowcolor[HTML]{EFEFEF} 
X-ViT-B & 82M & 45.8 & 41.2 \\ \hline
\end{tabular}
\caption{\textbf{Object detection performance on the COCO val2017.}}\label{tb:coco}
\end{table}

\section{Experiments} \label{sec:experiments}
\subsection{Image Classification}
\textbf{Dataset.} For the image classification task, we trained our models using the ImageNet1k\cite{deng2009imagenet} dataset from scratch.

\textbf{Implementation details.} Our setup was almost the same as that of DeiT\cite{touvron2020training}. However, we optimized some hyperparameters according to the model size. The learning rate was scaled per the 512 batch size following the linear scaling rule\cite{you2017large} and linearly warmed up for the first 5 epochs. We trained our model for 400 epochs using the AdamW optimizer\cite{loshchilov2017decoupled} and cosine scheduler. For data augmentation, CutMix\cite{yun2019cutmix} and RandAugment\cite{cubuk2020randaugment} was used. We applied a stronger augmentation in larger models.

\textbf{Fine-tune at higher resolution.} Instead of training from scratch again, we fine-tuned X-ViT-M and X-ViT-B at a higher resolution. Our models achieved better performance in 0.1$\times$ training time compared to learning from scratch.

\textbf{Comparison with state-of-the-art models.} We experimented with four models that used the same architectural design schemes as DeiT\cite{touvron2020training}. (See Table \ref{tb:imagenet_comparison}.) As summarized in Figure \ref{fig:params_acc_comparison}, all our models showed better performance and parameter efficiency than most of the concurrent transformer-based models.

\begin{figure}[t!]
\begin{center}
\includegraphics[width=\linewidth]{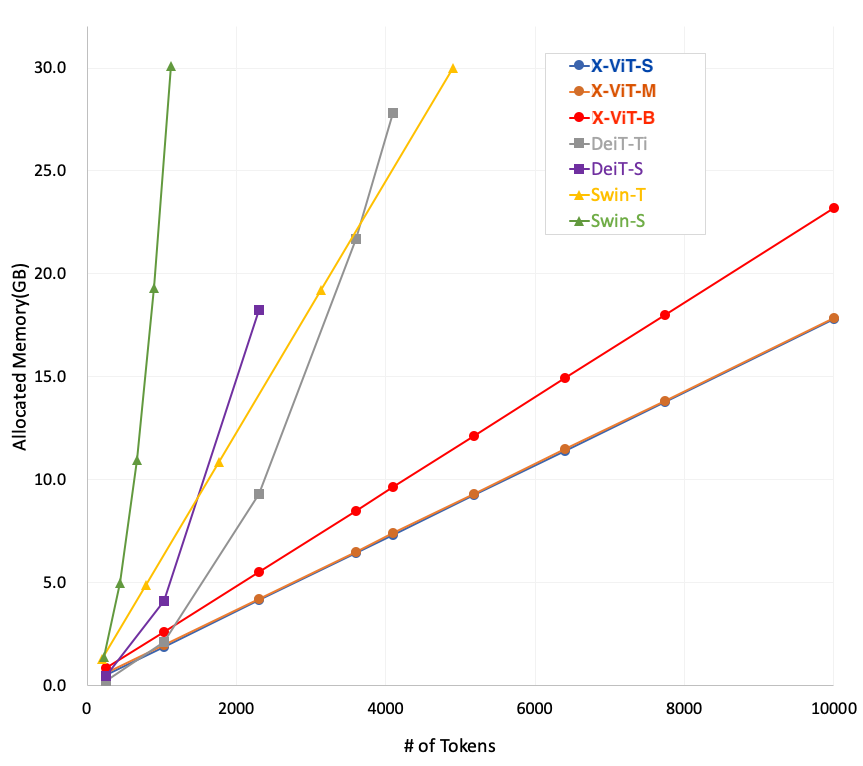}
\end{center}
\caption{\textbf{Allocated memory vs. \# of tokens.} To check the linearity of our models empirically, we measured the maximum value of allocated GPU memory on different resolutions. For a batch size of 64, the memory consumption of our models shows linearity with the number of tokens. Moreover, our models require significantly less memory than the other models.}
\label{fig:memory_consumption}
\end{figure}

\subsection{Object Detection with Mask R-CNN}
\textbf{Implementation details.} Our models were trained and evaluated on the COCO benchmark dataset\cite{lin2014microsoft} for the object detection task. We used our models as the backbone and mask R-CNN\cite{he2017mask} as the detector heads. Our training setups and hyperparameters follow that of DETR\cite{carion2020end}. All experiments were performed on a 3x schedule. The input resolution was fixed at $800\times1333$ for all the experiments.

\textbf{Evaluation on COCO dataset.} We compared CNNs\cite{he2016deep, xie2017aggregated} and ViT models on object detection and instance segmentation tasks. To make the comparison fair, the experimental environment was the same for all the results. All models were pre-trained on the ImageNet1k dataset.

According to Table \ref{tb:coco}, our models significantly outperform the CNN-based models and achieve higher or more competitive results than do state-of-the-art vision transformers. Notably, Swin transformer\cite{liu2021swin} models showed better results in the overall regime. Their architectural strategy is better optimized for dense prediction tasks, while that of our models is not.

\iffalse
The modules in XCiT\cite{el2021xcit} models have a structure similar to that of the X-ViT modules except for the SA scheme. Hence, they show a mAP curve similar to that of X-ViT according to the model capacity. They perform slightly better than our model in a similar design space. It is possible that the XCiT models have a larger embedding space $d$.

Swin transformer\cite{liu2021swin} models showed better results in the overall regime. We infer that this is because their architectural strategy is better optimized for dense prediction tasks. Notably, X-ViT-B performs slightly worse than X-ViT-M on the bounding box detection task, but slightly better in detecting smaller bounding boxes and overall instance segmentation scores.
\fi

\begin{figure}[t!]
\begin{center}
\includegraphics[width=\linewidth]{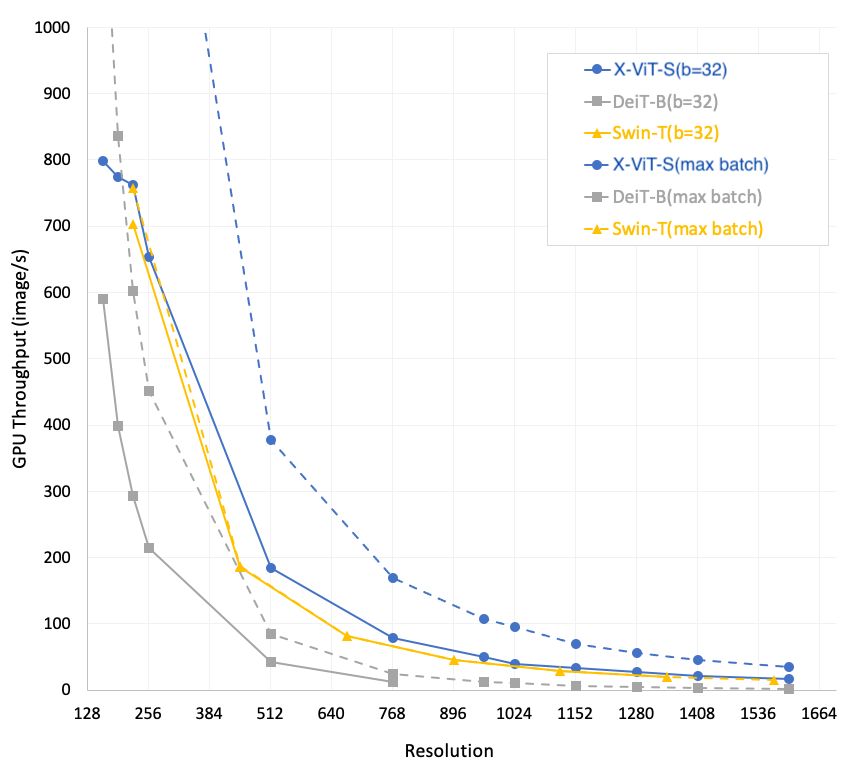}
\end{center}
\caption{\textbf{GPU throughput according to the input resolution.} Note that the scale of throughput axis is $\log _4$ scale. 'max batch' means throughput measured on maximum available batch size.}
\label{fig:resolution_throuput_comparison}
\end{figure}

\subsection{Measuring Computational Efficiency}
We measured the various computational resources required for the inference. All measurements were performed on a single V100 GPU with 32GB of VRAM.

\textbf{Memory efficiency.} According to Figure \ref{fig:memory_consumption}, we determined that our models consumed much less memory for larger resolutions compared to DeiT \cite{touvron2020training} and Swin Models\cite{liu2021swin}. Our model can process up to a 4$\times$ batch size compared with other models showing similar performance.

\textbf{GPU throughput.} Figure \ref{fig:resolution_throuput_comparison} shows that our model is faster than other models showing similar performance. And the GPU throughput of our model decreases more slowly compared to other models as input resolution increases.

\section{Conclusion}
In this paper, we proposed a simple method that ensures linear complexity for SA without
loss of performance. By replacing the softmax function, we removed the quadratic operation using the associative law of matrix multiplication. This type of factorization has typically caused performance degradation in earlier studies. The X-ViT models outperformed most of the existing state-of-the-art transformer-based and CNN-based models for image classification. We have shown that our models can also be deployed well for general purposes. Our X-ViT models show performance on dense prediction tasks that are competitive with or better than earlier models. With more optimized structures for dense prediction, we expect our models to become more efficient and perform better.

% \clearpage
% {\small
\bibliographystyle{ieee_fullname}
\bibliography{egbib}
% }
\clearpage
\end{document}